\documentclass{article}

\usepackage{microtype}
\usepackage{graphicx}
\usepackage{booktabs} 

\usepackage[ruled]{algorithm2e}
\usepackage{listings}
\usepackage{amsmath}
\usepackage{amssymb}
\usepackage{url}

\usepackage{listings}
\usepackage{subcaption}
\usepackage{mwe}
\usepackage{natbib}
\usepackage{xcolor}
\usepackage{multicol}

\usepackage[accepted]{mlsys2021}

\begin{document}

\twocolumn[
\mlsystitle{Accelerating SLIDE Deep Learning on Modern CPUs: Vectorization, Quantizations, Memory Optimizations, and More}



\mlsyssetsymbol{equal}{*}

\begin{mlsysauthorlist}
\mlsysauthor{Shabnam Daghaghi}{rice}
\mlsysauthor{Nicholas Meisburger}{rice}
\mlsysauthor{Mengnan Zhao}{rice}
\mlsysauthor{Yong Wu}{ant}
\mlsysauthor{Sameh Gobriel}{intel}
\mlsysauthor{Charlie Tai}{intel}
\mlsysauthor{Anshumali Shrivastava}{rice}
\end{mlsysauthorlist}

\mlsysaffiliation{rice}{Rice University}
\mlsysaffiliation{intel}{Intel Corporation}
\mlsysaffiliation{ant}{Ant Group (work was done while at Intel)}

\mlsyscorrespondingauthor{Shabnam Daghaghi}{shabnam.daghaghi@rice.edu}

\mlsyskeywords{Machine Learning, MLSys}

\vskip 0.3in

\begin{abstract}
Deep learning implementations on CPUs (Central Processing Units) are gaining more traction. Enhanced AI capabilities on commodity x86 architectures are commercially appealing due to the reuse of existing hardware and virtualization ease. A notable work in this direction is the SLIDE system. SLIDE is a C++ implementation of a sparse hash table based back-propagation, which was shown to be significantly faster than GPUs in training hundreds of million parameter neural models. In this paper, we argue that SLIDE's current implementation is sub-optimal and does not exploit several opportunities available in modern CPUs. In particular, we show how SLIDE's computations allow for a unique possibility of vectorization via AVX (Advanced Vector Extensions)-512.
Furthermore, we highlight opportunities for different kinds of memory optimization and quantizations. Combining all of them, we obtain up to 7x speedup in the computations on the same hardware. 
 Our experiments are focused on large (hundreds of millions of parameters) recommendation and NLP models. Our work highlights several novel perspectives and opportunities for implementing randomized algorithms for deep learning on modern CPUs. We provide the code and benchmark scripts at \url{https://github.com/RUSH-LAB/SLIDE}
\end{abstract}
]
\printAffiliationsAndNotice{}

\section{Introduction}

The recent surge in deep learning demand associated with an impressive performance on specialized hardware and GPUs (Graphical Processing Units) forced the data processing community to start considering newer architectures in their pipelines. Currently, specialized hardware and co-processors, including GPUs for accelerating tensor operations, are expensive additions. They are hard to virtualize~\cite{hong2017gpu} due to their technical architectures and instructions. Furthermore, they only provide benefits over CPUs for specialized forms of structured computations. For example, most modern GPUs and TPUs are only beneficial for operations that can be mapped into dense tensor operations. As a result, the true benefits of AI is only available to companies that can invest in massive infrastructural changes.

{\bf The Need for AI on CPUs:} Owing to their significance, deep learning implementations on CPUs (Central Processing Units) or x86 architectures have gained traction. Mainframe computers primarily consist of x86 architectures and are ubiquitous in the industry. The community has invested decades of efforts in virtualization~\cite{adams2006comparison} over x86 architectures, unifying a variety of platforms. If we can design systems that can make commodity CPUs at par with specialized processors like GPUs, it would be a big step towards AI's democratization.

{\bf CPUs have Large Memory for Large Networks:} Deep Learning architectures are exploding in sizes to assimilate the information present in large datasets. In 2017, Google introduced a 137-billion parameter network~\cite{shazeer2017outrageously}. The popular GPT-3~\cite{brown2020language} is a 175-billion parameter network. These networks require around a terabyte (TB) of memory to store the parameters. The state of practice requires hundreds or thousands of specialized processors, with only about 32-64 gigabytes of main memory, to train these outrageously large networks. It is not surprising that data moving cost itself dominates the training time. On the other hand, it is not difficult to assemble a single CPU machine with 3 Terabytes (or higher) of main memory~\cite{Angelini:2020,Zhang:2017}. Thus, if we can scale up deep learning computations on x86 architectures, we could, in theory, train these outrageously large models on a single machine in the near future.  

{\bf The Role of Sparsity: Dense Tensor Operation is Wasteful}
The current practice of training neural networks is to process and update all the parameters. The training process results in dense tensor operations, which are favorable for specialized hardware but results in unnecessary energy consumption. Recent work has shown that for extensively large networks, a very sparse update depending on the input is sufficient~\cite{ba2013adaptive}. It is quite wasteful to update hundreds of millions of parameters based on mistakes made on a handful of data instances. Exploiting this sparsity results in significantly cheaper alternatives. However, with the sparsity, the real advantage of specialized hardware over CPUs becomes less clear.  

{\bf SLIDE System on CPUs:}
Recently~\cite{spring2017scalable} showed that by leveraging probabilistic hash tables, we could efficiently identify very sparse ``active" sets of the parameters. Processing and updating only those parameters results in order of magnitude fewer computations. The idea opens the door for efficient and sparse gradient descent via approximate query processing instead of matrix multiplication. Approximate query processing already sits on decades of research from the systems and database community on x86 hardware. Furthermore, due to randomness and extreme sparsity of selected parameters for each data instance, we can parallelize gradient updates, thereby paving the way for vast asynchronous data parallelism. 

SLIDE system~\cite{chen2019slide}, combined these ideas into a C++  implementation for general CPUs with fewer multi-core parallelism. It was demonstrated that SLIDE, on a modest CPU, can be significantly faster in training a 200-million-parameter neural network, in terms of wall clock time, than the optimized TensorFlow implementation on NVIDIA V100 GPU. The most exciting part was that SLIDE was faster to reach any accuracy level than the competing baselines.  The focus of this paper will be on optimizing the existing SLIDE implementation for x86 architecture. 

{\bf Modern CPU: Opportunities and Challenges:} SLIDE implementation did demonstrate the power of smart algorithms on CPUs, but they only leveraged naive OpenMP parallelism. They did not exploit all the advantages modern CPUs have to offer. In particular, modern CPUs have evolved significantly in cache sizes and memory hierarchy. Vectorization is another component that is readily available in new CPUs for bulk (Single Instruction, Multiple Data - SIMD) operations. Recent CPU upgrades to AVX-512 instead of AVX-256, which allows processing 512 bits in a single CPU instruction, almost doubling the bulk processing capabilities. With multi-threading, modern CPUs can support significantly more parallel operations. Furthermore, Intel has recently started offering support for bfloat16 in their CPUs, allowing for faster low precision arithmetic. 

However, taking advantage of all these CPU functionalities are far from trivial. SLIDE is a very randomized workload where the memory access pattern is unpredictable and different in every data batch. As a result, leveraging cache hierarchy, advanced vectorization, etc. would require careful reconsideration of the SLIDE system. 

{\bf Our Contributions:} In this paper, we revisit the SLIDE system on two modern Intel CPUs in the hope of understanding the true potential of CPUs for training large deep learning models. We enable SLIDE to take advantage of vectorization, quantizations, and several memory optimizations in modern CPUs.  Our optimization effort leads anywhere from 2-7x training time speedup on the same hardware compared with the non-optimized SLIDE. We provide the code for both Optimized and Naive SLIDE and benchmark scripts for reproducibility \footnote{https://github.com/RUSH-LAB/SLIDE}.

We hope that this work will inspire several more research in novel algorithms and their optimized CPU implementations for large scale deep learning workloads. 


\section{Background: Sub-Linear Deep Learning Engine (SLIDE)}

SLIDE (Sub-LInear Deep Learning Engine) is a C++ OpenMP based system which combines smart hashing randomized algorithms with modest multi-core parallelism on CPU \cite{chen2019slide}. SLIDE's idea follows the line of adaptive sparsity/dropout \cite{blanc2018adaptive,ba2013adaptive,spring2017scalable} where a neural network can be accurately trained by only updating a small fraction of the neurons based on their activations. Specifically, SLIDE employs Locality Sensitive Hashing (LSH) to identify neurons during each update adaptively. It has been shown that SLIDE could outperform NVIDIA Tesla V100 GPU on several large datasets in terms of time-vs-accuracy comparison. 

\begin{figure*}
    \centering
    \includegraphics[scale=0.55]{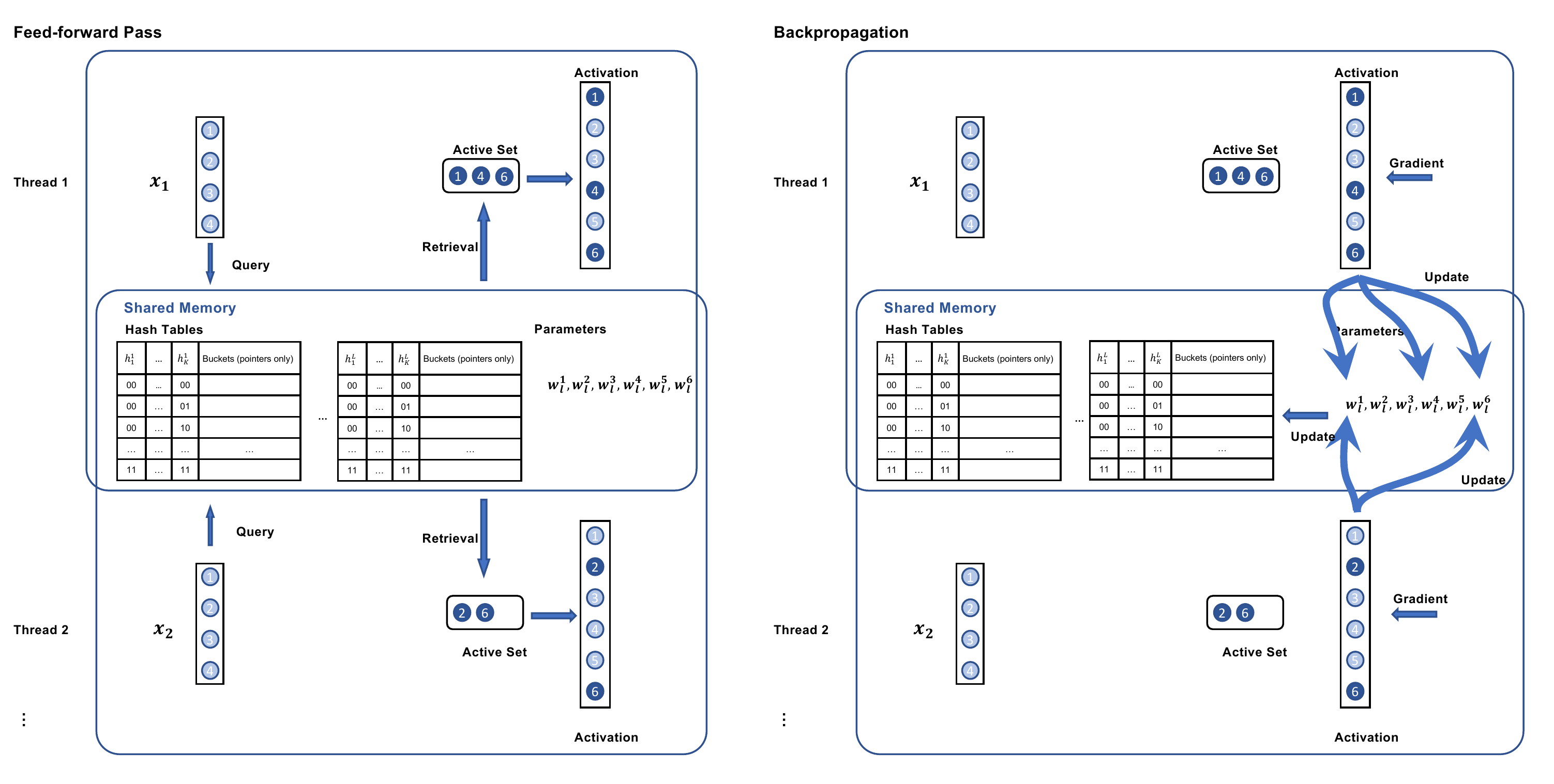}

    \caption{Illustration of feed-forward and backward pass. Two threads are processing the two data instances $x_1$ and $x_2$ in parallel with LSH hash tables and parameters in shared memory space. }\label{fig:feed-forward-and-backward}
\end{figure*}

The workflow of SLIDE is as the following:

\textbf{Initialization: } The weights of the network are randomly initialized in a standard way. Besides, each layer is initialized with a set of hash functions and corresponding hash tables. All the layer neurons are added (preprocessed) to the hash buckets based on their hash values. 

\textbf{Feed-forward Pass: } During the feed-forward pass, instead of computing the activation of all the neurons, we feed each layer's input into a hash function and query the hash table. The query process returns a small set of active nodes that are likely to have large activation on this input.  Only activations of the selected neurons are computed while other activations are treated as zero. On the left side of Fig.\ref{fig:feed-forward-and-backward} we show this sparse feed-forward pass for the simple case of two data instances $x_1$ and $x_2$. Different threads process the two data instances in parallel while the hash tables and weights of each layer are maintained in shared memory to be accessed. 

\textbf{Backpropagation and Hash Tables Update:} After computing the output and loss of the network, the error is back-propagated through the active neurons to perform parameter updates. For example, on the right-hand side of Fig.\ref{fig:feed-forward-and-backward}, only the weights of neurons 1, 4, and 6  for data $x_1$  and the weights of neurons 2 and 6 for data $x_2$ are updated. If a fraction $p$ of neurons in one layer is active, the fraction of weights that would be updated is only $p^2$, which is a significant reduction of computation when $p \ll 1$. Afterward, the hash tables are updated according to the new weights' values. If the neuron's buckets in the hash tables need to be updated, it will be deleted from the current bucket in the hash table and re-add it. We refer the interested readers to \cite{chen2019slide} for a detailed review of the algorithm.

{\bf HOGWILD Style Parallelism} One unique advantage of SLIDE is that every data instance in a batch selects a very sparse and a small set of neurons to process and update. This random and sparse updates open room for HOGWILD~\cite{recht2011hogwild} style updates where it is possible to process all the batch in parallel. SLIDE leverages multi-core functionality by performing asynchronous data parallel gradient descent.

\section{Target CPU Platforms: Intel  CLX and Intel CPX}\label{sec:CPXCLX}

In this paper, we target two different, recently introduced, Intel CPUs server to evaluate the SLIDE system and its effectiveness: {\bf Cooper Lake server (CPX)}~\cite{Intel:CPX} and {\bf Cascade Lake server (CLX)}~\cite{Intel:CLX}. Both of these processors support AVX-512 instructions. However, only the CPX machine supports bloat16. 

CPX is the new 3rd generation Intel Xeon scalable processor, which supports AVX512-based BF16 instructions. It has Intel  x86\_64 architecture with four 28-core CPU totaling 112 cores capable of running 224 threads in parallel with hyper-threading.  The $L3$ cache size is around 39MB, and $L2$ cache is about 1MB.

CLX is an older generation that does not support BF16 instructions. It has Intel  x86\_64 architecture with dual 24-core processors (Intel Xeon Platinum 8260L CPU @ 2.40GHz) totaling 48 cores. With hyper-threading, the maximum number of parallel threads that can be run reaches 96. The $L3$ cache size is 36MB, and $L2$ cache is about 1MB.

\section{Opportunities for Optimization in SLIDE}\label{sec:optimize_SLIDE}

The SLIDE C++ implementation is open sourced codebase\footnote{https://github.com/keroro824/HashingDeepLearning} which allowed us to  delve deeper into its inefficiencies. In this section, we describe a series of our efforts in exploiting the unique opportunities available on CPU to scaling up SLIDE.

\subsection{Memory Coalescing and cache utilization}
Most machine learning jobs are memory bound. They require reading continuous data in batches and then updating a large number of parameters stored in the main memory. Similar is the case for SLIDE implementation.

{\bf Understanding Memory Loading Mechanism in CPU:}  Main memory gets loaded via \emph{cache line chunks} - usually 64 or 128 bytes - through a hierarchy of caches: from DRAM to $L3$ to $L2$ and $L1$ and finally to the CPU registers. As a result, when a location in DRAM (CPU main memory) is accessed, a number of consecutive locations within the same cache line are loaded  into the cache. Furthermore, most processors automatically pre-fetch the next cache line even without a request in anticipation that this cache line will be processed in the near future (In reality, different hardware prefetchers are present in the platform working in parallel, where each targets a different access pattern, however, sequential prefetchers are very common for many workloads). The pre-fetching is a feature to hide memory load latency and assumes a serial read.  For different caches, the same kind of auto-loading and auto-fetching behavior happens. Modern cache sizes have increased significantly, and we can have around 35MB of space in $L3$ cache shared with all the processors.  Reading from $L3$ cache is approximately 3x faster than from DRAM. Effectively, when a memory location is requested, several memory locations contiguous to the requested address are available for significantly faster read automatically. 

When all threads execute a load instruction, the most favorable scenario would be for all threads to access consecutive global memory locations. In this case, the hardware could coalesce these memory accesses into single access to consecutive DRAM locations. Thus, most threads are loading from the cache directly, which is at least 3x as fast as loading from DRAM. 

Most randomization algorithms, including SLIDE, ignore this phenomenon because they assume that the memory access pattern is random enough to exploit the cache.  We found that SLIDE has major memory fragmentation, which increases the data access latency - the fragmented memory results in inefficient cache utilization on CPUs. There are two significant causes of memory fragmentation in SLIDE: 1) Data Memory Fragmentation and 2) Parameter Memory Fragmentation. 

{\bf Removing Data Memory Fragmentation} SLIDE works with sparse datasets and sparse activations. Due to the variable number of non-zeros, a commonly adopted choice is to use different arrays (or vectors) to store various data instances. However, this creates unnecessary caches misses. SLIDE uses HOGWILD style parallelism~\cite{recht2011hogwild} in which hundreds of data instances are processed simultaneously by several different threads. With the standard implementation of the sparse format in SLIDE, every thread will access data from a wildly different DRAM memory location, which does not take advantage of cache, especially the $L3$ (Level 3) cache shared across all the threads.  

Instead of fragmenting every data instance, we create one long contiguous vector that consecutively stores non-zero indices and values of all the data instances in a batch. To keep track of variable numbers of non-zeros, we keep another vector of offsets. These offset values will  be used to index the starting location of any data instance directly. In practice, we process hundreds to thousands of data instances using hundreds of threads in parallel. As a result, When the first data instance is sought from the DRAM, most following consecutive data elements in the batch are loaded/pre-fetched in the cache, readily available for other threads reducing memory latency. 

{\bf Removing Parameter Memory Fragmentation:} In the SLIDE algorithm, every neuron in each layer is activated independently, and the identity of active neurons is dependent on the input. Thus,  every input instance selects the neurons dynamically by querying the hash tables. The very high sparsity, which is the key algorithmic benefit of SLIDE, generally ensures that two different threads select a near disjoint set of neurons. However,  in a heavy multi-core environment, hundreds of data batches are processed in parallel, resulting in a large set of processed neurons. As a result, several adjacent neurons are likely required in each batch. This phenomenon creates another opportunity for memory coalescing. 

It should be noted that the exact same neuron activated in multiple threads (i.e. shared across threads) does not lead to memory coalescing because it is anyway available when needed. This event is nevertheless rare in SLIDE, which is the reason why HOGWILD style updates work. When processing a batch, if neuron $\nu$ is selected by a given thread, even though we only use a sparse subset of its weight vector, other threads will likely use nearby subsets or other weights that are located nearby in memory. Thus, while fetching neuron $\nu$ from DRAM, we will have the nearby neurons in the cache readily available for other threads to consume increasing cache utilization. Since SLIDE is a randomized algorithm, it is impossible to predict the set of neurons selected in a batch. However, even with a random selection of neurons, a significant number of neurons in nearby locations will be chosen when the batch size is reasonably large.

Therefore, we ensure that even though neurons are independent entities, neuron weights in the same layer are located contiguously in the main memory increasing cache efficiency. We reserve a big chunk of contiguous memory that contains different neuron weights in a given layer. 
\subsubsection{Hyper-threading for HOGWILD Style Updates}

Hyper-Threading allows the processor to virtually double its number of cores and execute twice as many threads simultaneously, jumping back and forth between threads to execute instructions for one whenever the other is stalled. Since SLIDE gradient update is HOGWILD style data-parallel and asynchronous, hyperthreading provides a significant boost. Due to the randomized nature of SLIDE memory access, threads do get stalled due to cache miss. At that point, hyper-threading will be favorable because another thread can utilize the wait time to process a different data instance.

\subsection{Vectorization with AVX-512}
Moore's Law ensures performance improvement across generations of hardware by consistently providing increased parallelism through additional cores, and wider SIMD (Single instruction, multiple data) registers~\cite{Intel:SIMD}. AVX (Advanced Vector Extensions)-512 are newer SIMD instructions to 512-bit for x86 set architecture. 

To execute CPU instructions such as adds, multiplies, etc. the values operated on must be loaded into registers. Overtime registers have grown in size, allowing for either larger values to be stored and operated on and allowing for multiple values to be stored in a single wider register. It is this later capability that has been extended to allow for vectorization. Most CPUs today contain 64-bit registers, but select newer processors contain up to 512-bit registers \cite{512bit_register}. The system uses AVX-512 instructions, which are instructions that allow for a single operation, such as an add, to be applied to the "vector" of values that reside in a 512 bit register \cite{Reinders:2017}. For example, suppose we wish to do a pairwise addition on two arrays, each containing 16 32-bit integers. In that case, we can load each of these arrays into an 512-bit register, and add them together in a single operation, and write the resulting value back to a new array. Figure \ref{fig:avx_code} illustrates an example.

\begin{figure}
\begin{scriptsize}
    \centering
\begin{lstlisting}[language=c++,breaklines]
int main() {
    int x[16] = {1,2,3,...,15,16};
    int y[16] = {2,4,6,...,30,32};
    int s[16]; 
    // Load the arrays in 512 registers
    __m512i xr = _mm512_load_epi32(&(x[0]));
    __m512i yr = _mm512_load_epi32(&(y[0]));
    // Compute sum using AVX
    __m512i sum = _mm512_add_epi32(xr, yr);
    // Result in the final array.
    _mm512_store_epi32(&(s[0]), sum);
    return 0;}
\end{lstlisting}
\end{scriptsize}
    \caption{Code for adding 16 numbers in one SIMD Instruction.}
    \label{fig:avx_code}
\end{figure}

\subsection{AVX-512 in SLIDE}

\subsubsection{Parameter Updates with ADAM}

Vectorizing ADAM parameter updates is very straightforward. Each item in the weight matrix is updated according to the standard vector formula. 

\begin{figure}
    \centering
    \includegraphics[scale=0.65]{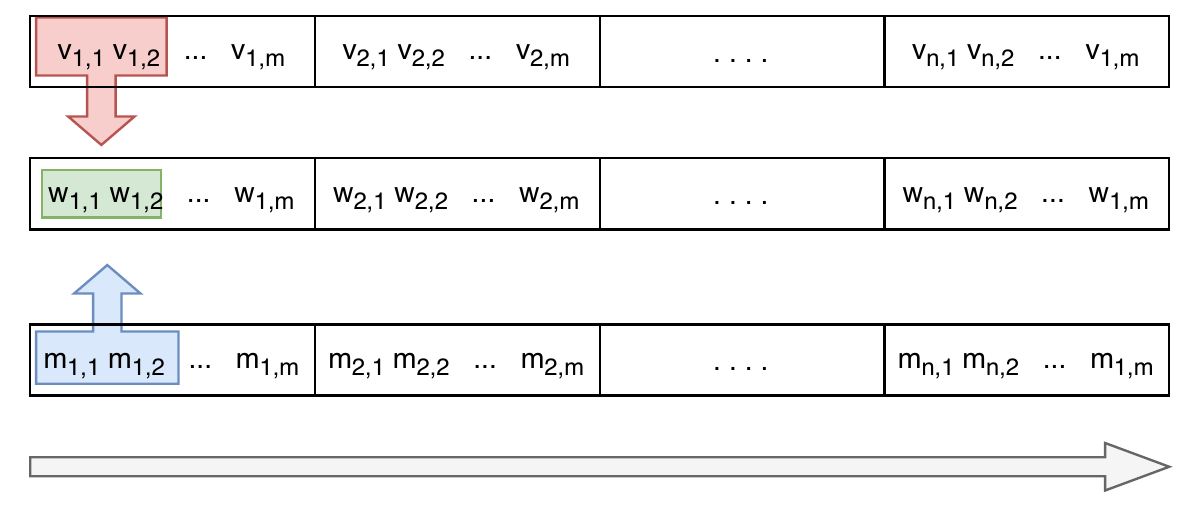}
    \caption{Vectorization of ADAM Parameter Update}
    \label{fig:avx_adam}
    \vspace{-0.2in}
\end{figure}

Since all matrices are represented as contiguous blocks of memory, we can simplify the process from 2D to a 1D loop. Essentially we take a 1D loop and iterate over the arrays in memory by increments of either 16 or 32 (assuming the weights are represented as 32-bit or 16-bit floats). Figure~\ref{fig:avx_adam} illustrates how we vectorize velocity, momentum, and gradient updates. 

\subsubsection{Vectorizing Sparse-Dense and Dense-Sparse Operations in SLIDE}
The workflow of SLIDE is very peculiar and opens up room for a specific kind of vectorization. We start by defining a few concepts. 

Let's say that layer $\ell$ with dimension $n$, connected to a previous layer with dimension $m$, has a weight matrix $\mathbf{W_\ell} \in \mathbb{R}^{n\times m}$. Where row $i$ represents the weight vector for neuron $i$.

\textbf{Definition:} We say that a matrix is in \textbf{Row-Major Order} if foreach weight vector in the layer, the components of the weight vector are adjacent to each other in memory, that is the matrix is in the form $W =[\mathbf{w_{1,1}}, \mathbf{w_{1,2}} \dots \mathbf{w_{1,m}}, \dots, \mathbf{w_{n,1}}, \mathbf{w_{n,2}}\dots \mathbf{w_{n,m}}]$.

\textbf{Definition:} We say that a matrix is in \textbf{Column-Major Order} if foreach component of each weight vector, components corresponding to the same dimension are adjacent to each other in memory. Thus the matrix is represented as follows $W = [\mathbf{w_{1,1}}, \mathbf{w_{2,1}}\dots \mathbf{w_{n,1}}, ..., \mathbf{w_{1,m}}, \mathbf{w_{2,m}}\dots \mathbf{w_{n,m}}]$.

If a vector, $\mathbf{x}$ is \textbf{sparse}, meaning that most of the components of $\mathbf{x}$ is zero, we represent $\mathbf{x}$ as a set of (index, value) pairs. $\mathbf{x} := \{ (i, v_i) \in \mathbb{Z}^+ \times \mathbb{R} : 1 \leq i \leq m \land v_i \neq 0 \} $. Otherwise, if $\mathbf{x}$ is \textbf{dense} then we represent $\mathbf{x}$ a series of continuous values $\mathbf{x} := [ v_1, v_2, ... v_m ] $.

\textbf{Lemma 1:} If a weight matrix $\mathbf{W}$ is in column-major order, then it is in row-major order for $\mathbf{W^T}$ and similarly, if a weight matrix is in row-major order for $\mathbf{W}$ then it is in column-major order for $\mathbf{W^T}$

SLIDE is most appealing for wide layers. Typically, the input instance to the wide layers is dense during the forward pass. However, the output is very sparse because very few neurons get activated. So if we think of $\mathbf{x}$ as input and $\mathbf{W}$ as weight vectors, depending on the activation, we are interested in only a few sparse entries of $\mathbf{Wx}$. We have the opposite in the gradient descent step; we have a very sparse output gradient $\nabla \mathbf{y}$, which gets multiplied by $\mathbf{W^T}$ resulting in a dense gradient for the next layer to propagate further. Effectively, both the forward pass contains a series of multiplications $\mathbf{y} = \mathbf{W} \mathbf{x}$, and the backward pass contains a series of multiplications in the form $\nabla \mathbf{W} = \nabla \mathbf{y} \cdot \mathbf{x^T}$, and $\nabla \mathbf{x} = \nabla \mathbf{y^T} \mathbf{W}$, or equivalently, $\nabla \mathbf{x} = \mathbf{W^T} \nabla \mathbf{y}$. 

It turns out that we can perform both of these operations efficiently. First, let us start with another easy lemma. 

\textbf{Lemma 2:} We can vectorize $y = WX$ computations if one of the following conditions is meet:
\begin{enumerate}\vspace{-0.1in}
    \item $\mathbf{x}$ is dense and $\mathbf{W}$ is in row-major order.\vspace{-0.1in} 
    \item $\mathbf{y}$ is dense and $\mathbf{W}$ is in column-major order.\vspace{-0.1in} 
\end{enumerate}

\begin{figure}[h]
    \centering
    \includegraphics[scale=0.55]{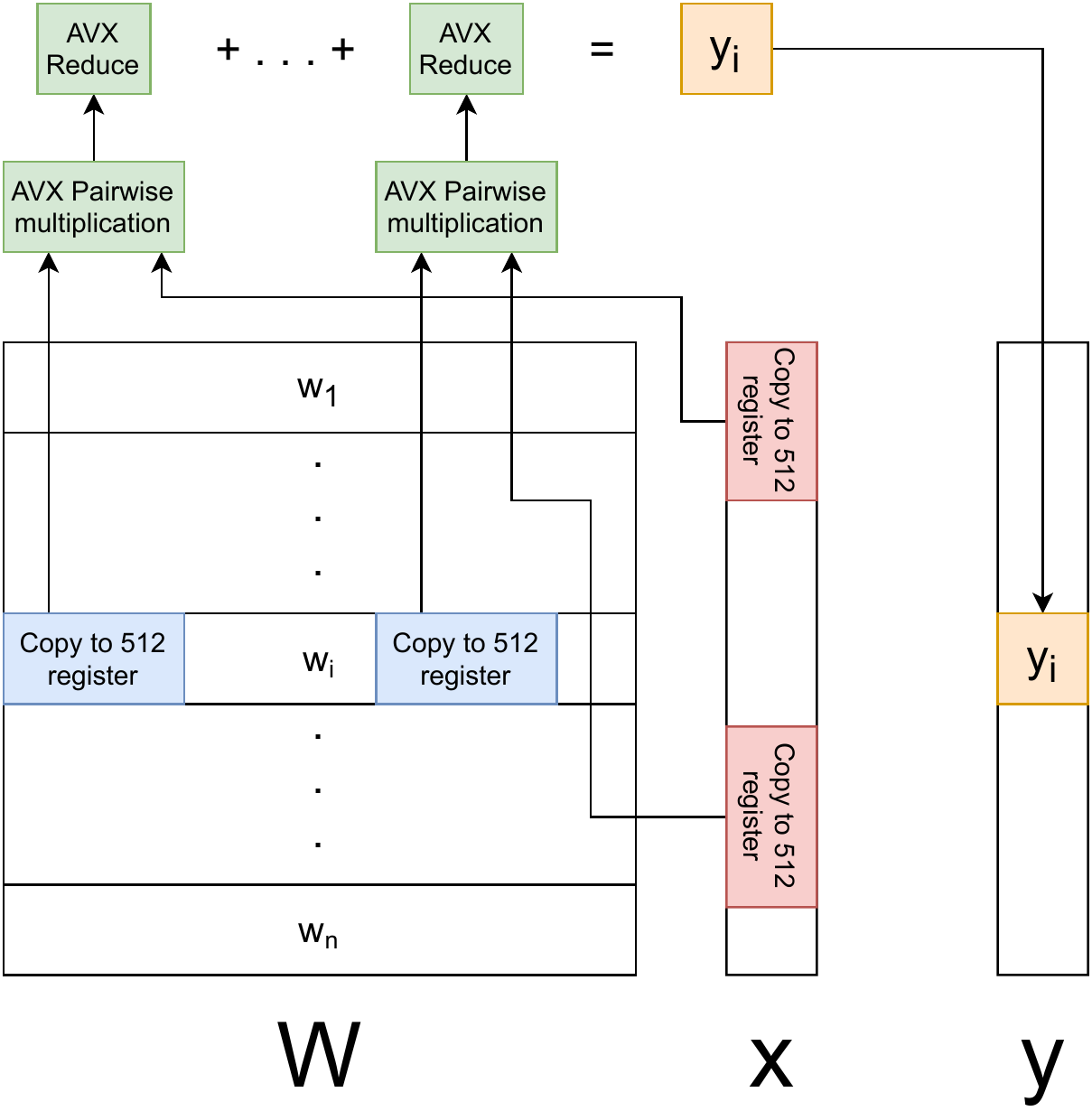}
    \caption{Matrix-vector multiplication with dense $\mathbf{x}$ and sparse or dense $\mathbf{y}.$}
    \label{fig:denseX}
\end{figure}

Let's examine case 1 first. Because $\mathbf{x}$ is dense, we can decompose the matrix-vector multiplication into a series of dense inner products. This approach is agnostic to whether or not $\mathbf{y}$ is dense or sparse; if it is dense, we will iterate over all the neurons weight vectors and compute their inner product with $\mathbf{x}$; if it is sparse we can do the same process for only the active neurons. To vectorize the inner product computation we can iterate over the 2 vectors, 16 or 32 elements at a time, compute their pairwise multiplication using AVX instructions, and use the AVX reduce sum instruction to combine these products into a single sum, which we will accumulate as we iterate over the vectors, yielding a single sum at the end. Figure~\ref{fig:denseX} illustrate the idea. Algorithm~\ref{alg:densex} provides the AVX-512. Because the matrix is in row-major order, the components of each weight vector are adjacent. They will maximize cache hits as we iterate over the vector and easily load the vectors' blocks into the 512-bit registers.

\begin{algorithm}
\caption{Matrix-vector multiplication with dense $\mathbf{x}$ and sparse or dense $\mathbf{y} = Wx$}
\label{alg:densex}
\KwIn{A dense vector $\mathbf{x}$, and matrix $\mathbf{W} \in \mathbb{R}^{n\times m}$ in row major order, and a set of active neurons $A$}
\hrulefill\\
\nl \For{$i \in A$}{
    \nl $s \gets 0$\\
    \nl \For{$j \gets 1;\ j \leq m;\ j \gets j+16$}{
        \nl $\mathbf{mm512}\ xr \gets \mathbf{mm512\_load}(\&(\mathbf{x} + j))$\\
        \nl $\mathbf{mm512}\ wr \gets \mathbf{mm512\_load}(\&(\mathbf{W} + i \cdot m + j))$\\
        \nl $\mathbf{mm512}\ sum = xr \cdot wr$\\
        \nl $s \gets s + \mathbf{mm512\_reduce}(sum)$\\
    }
    \nl $\mathbf{y.push\_back}(i, s)$\\
}
\nl \Return{$\mathbf{y}$}
\end{algorithm}

\begin{figure}
    \centering
    \includegraphics[scale=0.55]{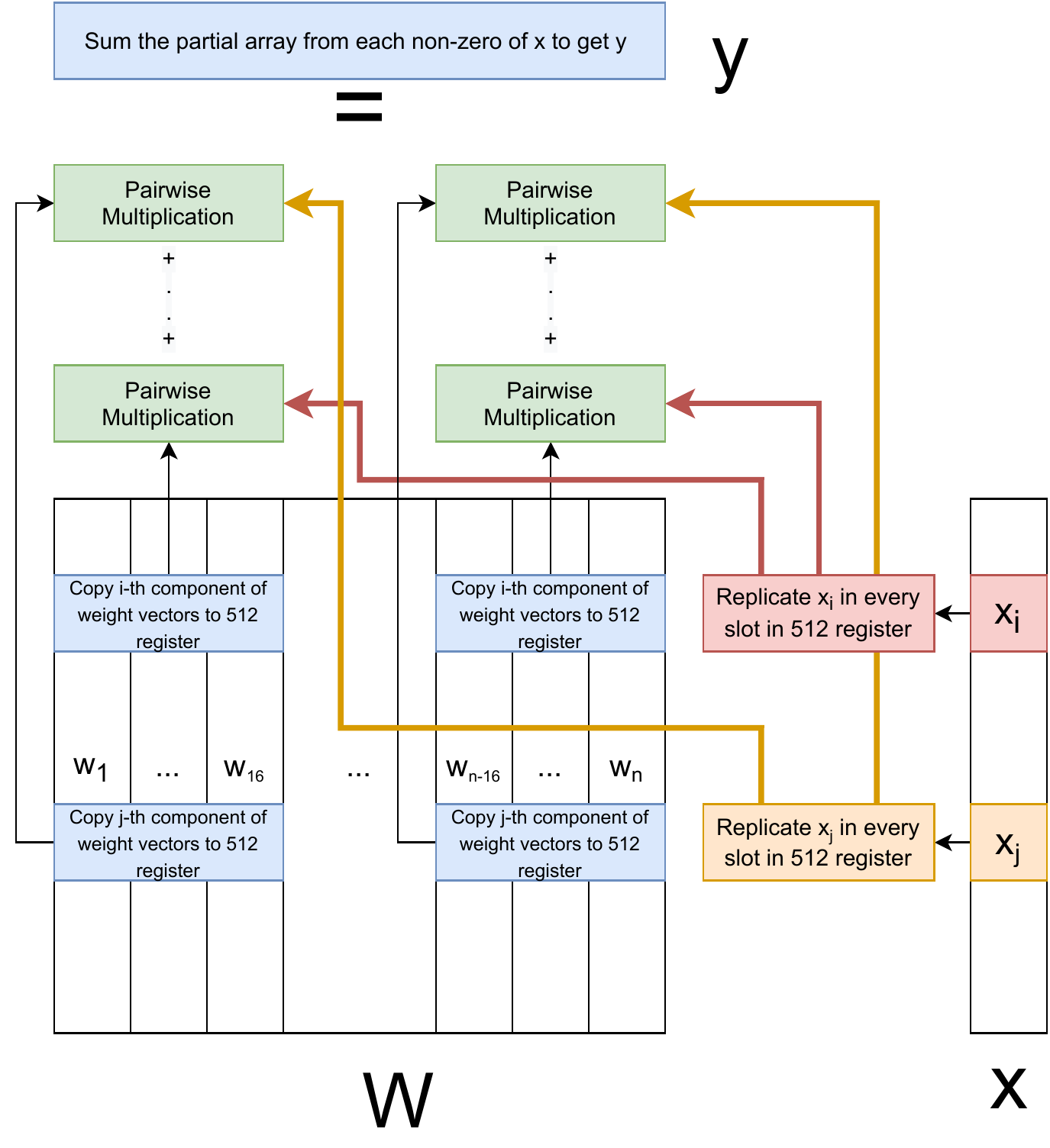}
    \caption{Matrix-vector multiplication with sparse or dense $\mathbf{x}$ and dense $\mathbf{y}$}
    \label{fig:sparseX}
\end{figure}

Now let's consider case 2, which is slightly more complicated. Because $\mathbf{x}$ is sparse, the non zero values in $\mathbf{x}$ are adjacent in memory, but the components of the weight vectors they correspond to are not, so the same approach to vectorizing the inner product cannot be applied. The solution here lies in the fact that $\mathbf{W}$ is in column-major order, meaning that the $i$th component of each weight vector is adjacent. The approach to vectorizing this multiplication is to iterate over the non-zero elements of $\mathbf{x}$. Foreach non zero of $\mathbf{x}$, $(i, v_i)$ fill one 512 bit register with $v_i$ repeated in every slot. Then iterate over the $i$th components of each of the weight vectors by increments of 16 or 32, loading them into a separate 512-bit register and computing the pairwise multiplication with the first register containing $v_i$. This results in a component of each of the activations for the subsequent layer. By summing these intermediate vectors for each non-zero term of $\mathbf{x}$, we get the final activation vector. This process works because $\mathbf{W}$ is in column-major order, so the $i$th component of each vector is adjacent. Figure~\ref{fig:sparseX} illustrates the idea, and the AVX-15 code is provided in Algorithm~\ref{alg:sparsex} for convenience.

\begin{algorithm}
\caption{Matrix-vector multiplication with sparse or dense $\mathbf{x}$ and dense $\mathbf{y}= Wx$}
\label{alg:sparsex}
\KwIn{A sparse vector $\mathbf{x}$, and matrix $\mathbf{W} \in \mathbb{R}^{n\times m}$ in column major order}
\hrulefill\\
\nl $\mathbf{y} \gets [\ \mathbf{y_i} = 0\ \forall i \in [1,n]\ ]$\\
\nl \For{$(j, v_j) \in \mathbf{x}$}{
    \nl \For{$i \gets 1;\ i \leq n;\ i \gets i+16$}{
        \nl $\mathbf{mm512}\ xr \gets \mathbf{mm512\_fill}(v_i)$\\
        \nl $\mathbf{mm512}\ wr \gets \mathbf{mm512\_load}(\&(\mathbf{W} + j \cdot n + i))$\\
        \nl $\mathbf{mm512}\ sum = xr \cdot wr$\\
        \nl $\mathbf{mm512}\ yr \gets \mathbf{mm512\_load}(\&(\mathbf{y} + i))$\\
        \nl $\mathbf{mm512\_store}(\&(\mathbf{y} + i), yr + sum)$
    }
}
\nl \Return{$\mathbf{y}$}
\end{algorithm}

\begin{table*}
    \centering
    \footnotesize
    \caption{Statistics of the datasets}
    \begin{tabular}{|c|c|c|c|c|c|c|} \hline
    	& Feature Dim & Feature Sparsity & Label Dim & Train Size & Test Size &  \# Model Parameters \\ \hline
    	Amazon-670K & 135909 & 0.055 \% & 670091 & 490449 & 153025  & 103 million \\ \hline
    	WikiLSHTC-325K & 1617899 & 0.0026 \% & 325056 & 1778351 & 587084 & 249 million \\ \hline
    	Text8 & 253855 & 0.0004 \% & 253855 & 13604165 & 3401042 & 101 million\\
    	\hline\end{tabular}
    	\label{table:data}
\end{table*}
We now come to the important issue: let's say that we are using algorithm \ref{alg:sparsex} on a sparse input vector $\mathbf{x}$ and a dense output vector $\mathbf{y}$, meaning our weight matrix is in column-major order. Thus,  in the backpropagation step, we are interested in multiplying $\mathbf{y}$ times $\mathbf{W}$. However, this means we are multiplying the matrix by a dense vector to get a sparse vector. It seems to contradict \textbf{Lemma 2}, as the matrix is in column-major order, but row-major order is required for algorithm \ref{alg:densex}. However this fact can be reconciled by observing that in the backpropagation step we are interested in computing $\nabla \mathbf{x} = \mathbf{W^T} \nabla \mathbf{y}$. By \textbf{Lemma 1}, we know that column-major order for $\mathbf{W}$ is row-major order for $\mathbf{W^T}$. Overall, our conditions for applying algorithm \ref{alg:densex} are met. Similarly, if we are applying algorithm \ref{alg:densex} in the forward direction and must therefore apply algorithm \ref{alg:sparsex} in the backpropagation phase, we can apply \textbf{Lemma 1} again to ensure that we satisfy the conditions of algorithm \ref{alg:sparsex}.

\subsubsection{Vectorizing Densified-Winner-Takes-All (DTWA)}

One important computational bottleneck in SLIDE is hash computations. SLIDE uses Locality Sensitive Hashing (LSH)~\cite{andonie2lsh} to sample high activation neurons. In particular, SLIDE uses recently proposed DTWA hashing which works nicely on sparse data. If we represent the data vector as a list of indices and values. DTWA operation computes a random hash map of every non-zero index in the list. Then the map is partitioned into a fixed number of bins. Finally, the index with the maximum coordinate value in each bin is the hash value.  See~\cite{chen2018densified} for details. 

To efficiently vectorize this LSH function. We pre-compute the random map of all the indices.  We then use the standard AVX-512 vectorized instruction for max operations in each bin after aggregating the locations falling in it. 

\subsection{BF16 Optimization}
\label{section:BF16Opt}
Brain floating-point format (BF16) uses 16 bits in computer memory to represent a floating-point number~\cite{kalamkar2019study}. It has been used widely in hardware accelerating machine learning. Compared with FP32, BF16 truncates the mantissa from 23 bits to 7 bits while preserving the exponent with 8 bits. We propose two modes utilizing BF16 to accelerate the training process. The first mode is to represent activation and weight both in BF16 format. The second mode represents activation in BF16 while updating parameters in FP32 to avoid degrading training quality while improving speed. Replacing 32-bit operations with  16-bit ones is likely to increase the computation speed. It also enhances the power of AVX-512 by doubling the number of operations for the same number of instructions.

\begin{figure*}[ht]
\begin{center}
\begin{multicols}{3}
    \includegraphics[width=1\linewidth]{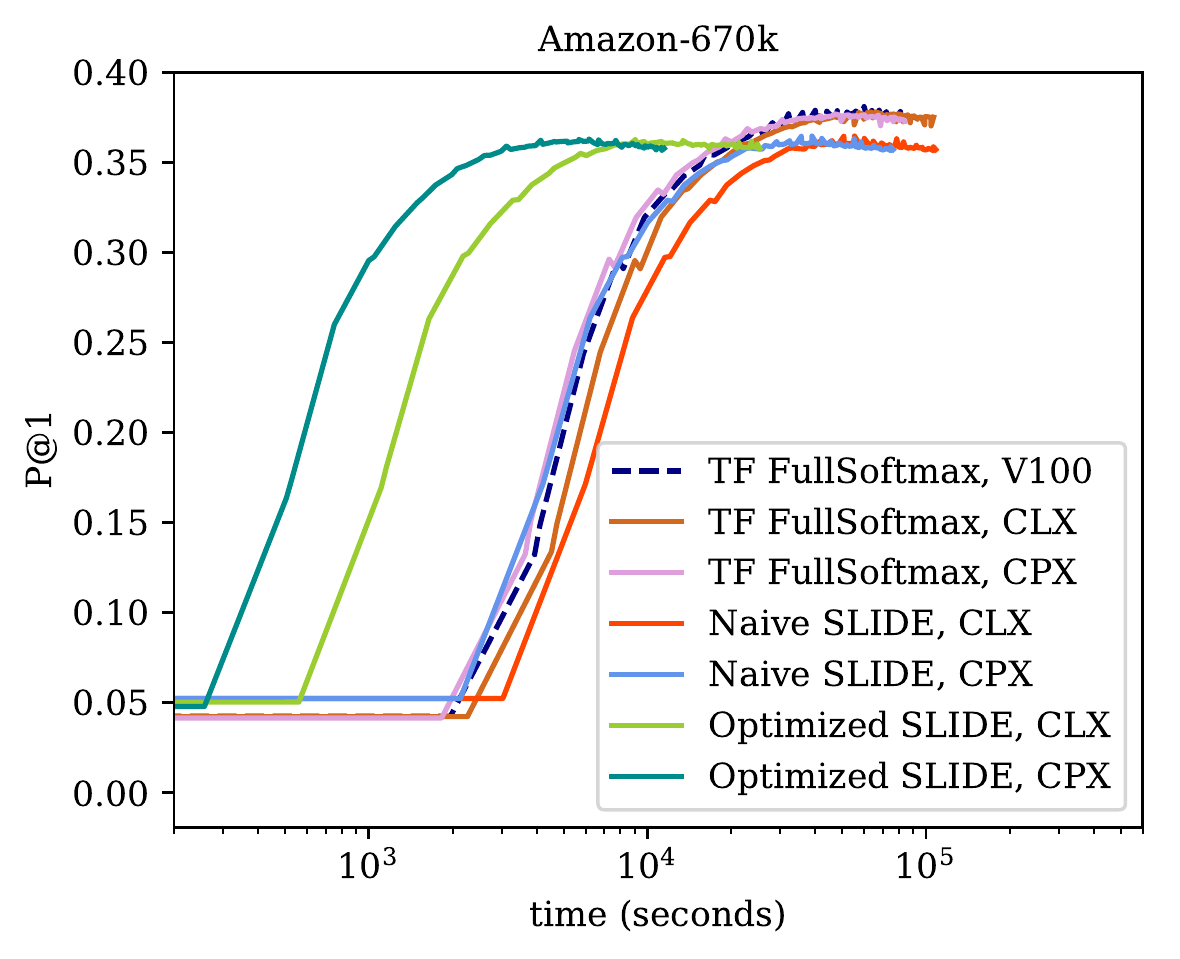}\par 
      \label{}
         \includegraphics[width=1\linewidth]{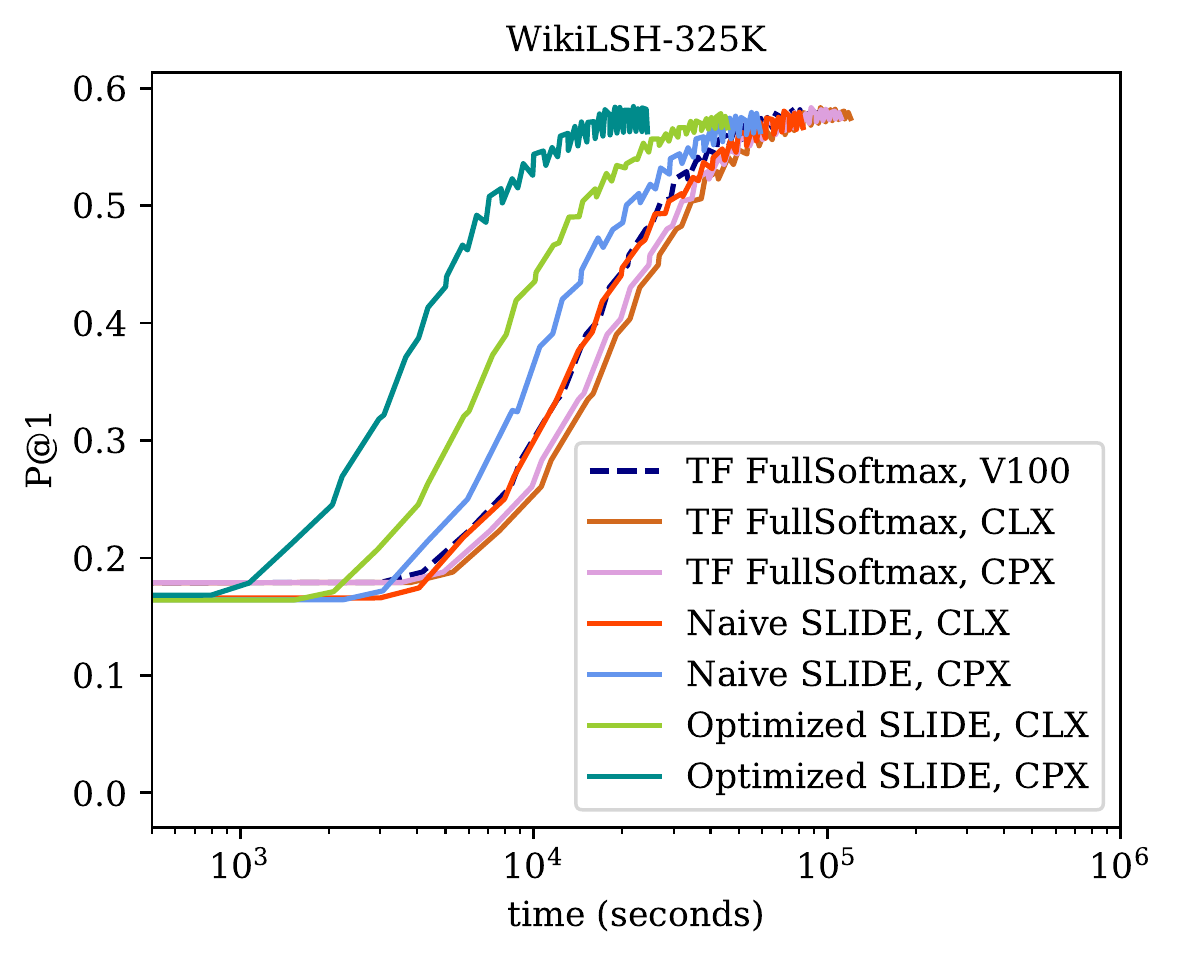}\par 
      \label{}

    \includegraphics[width=1\linewidth]{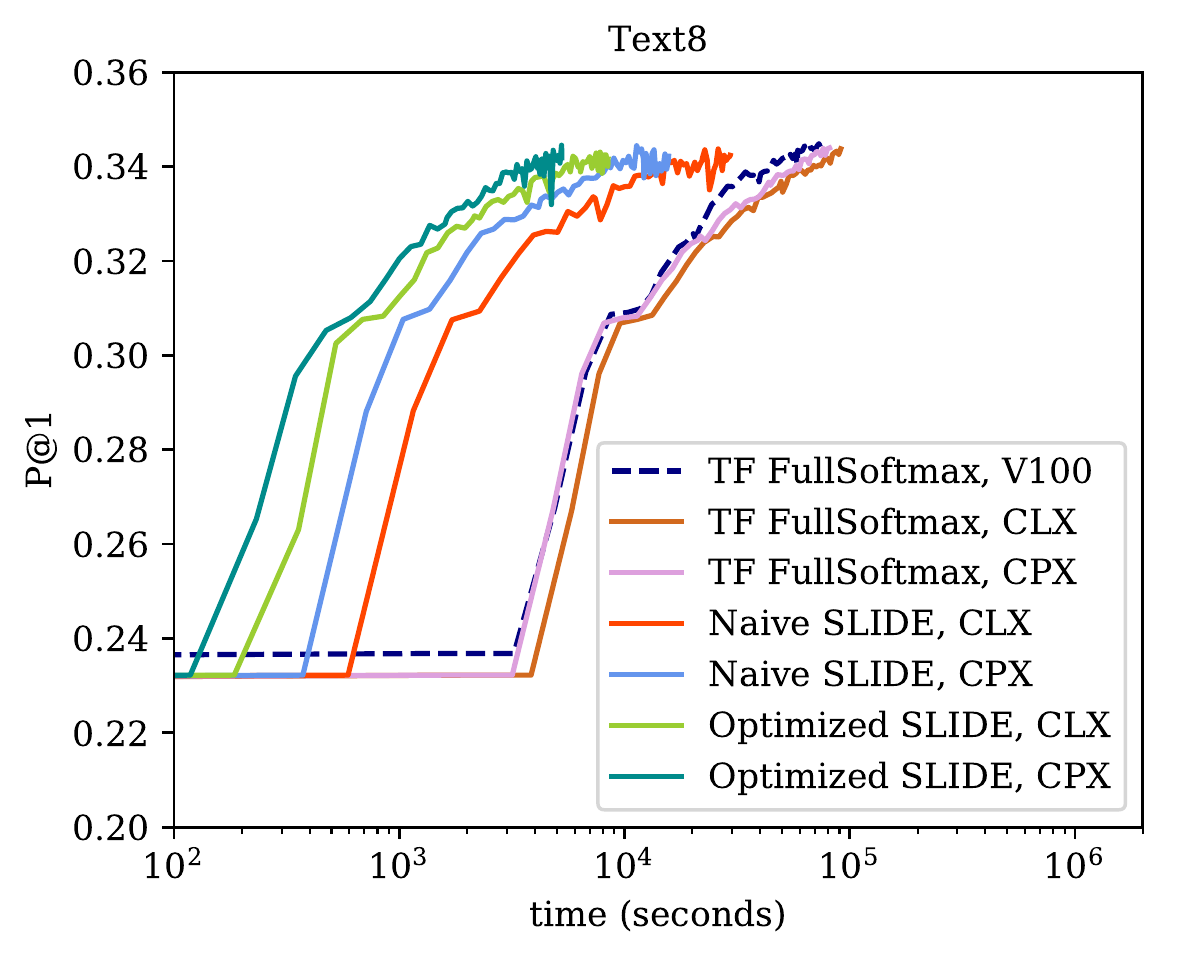}\par 
      \label{}

\end{multicols}
    \begin{multicols}{3}
    \includegraphics[width=1\linewidth]{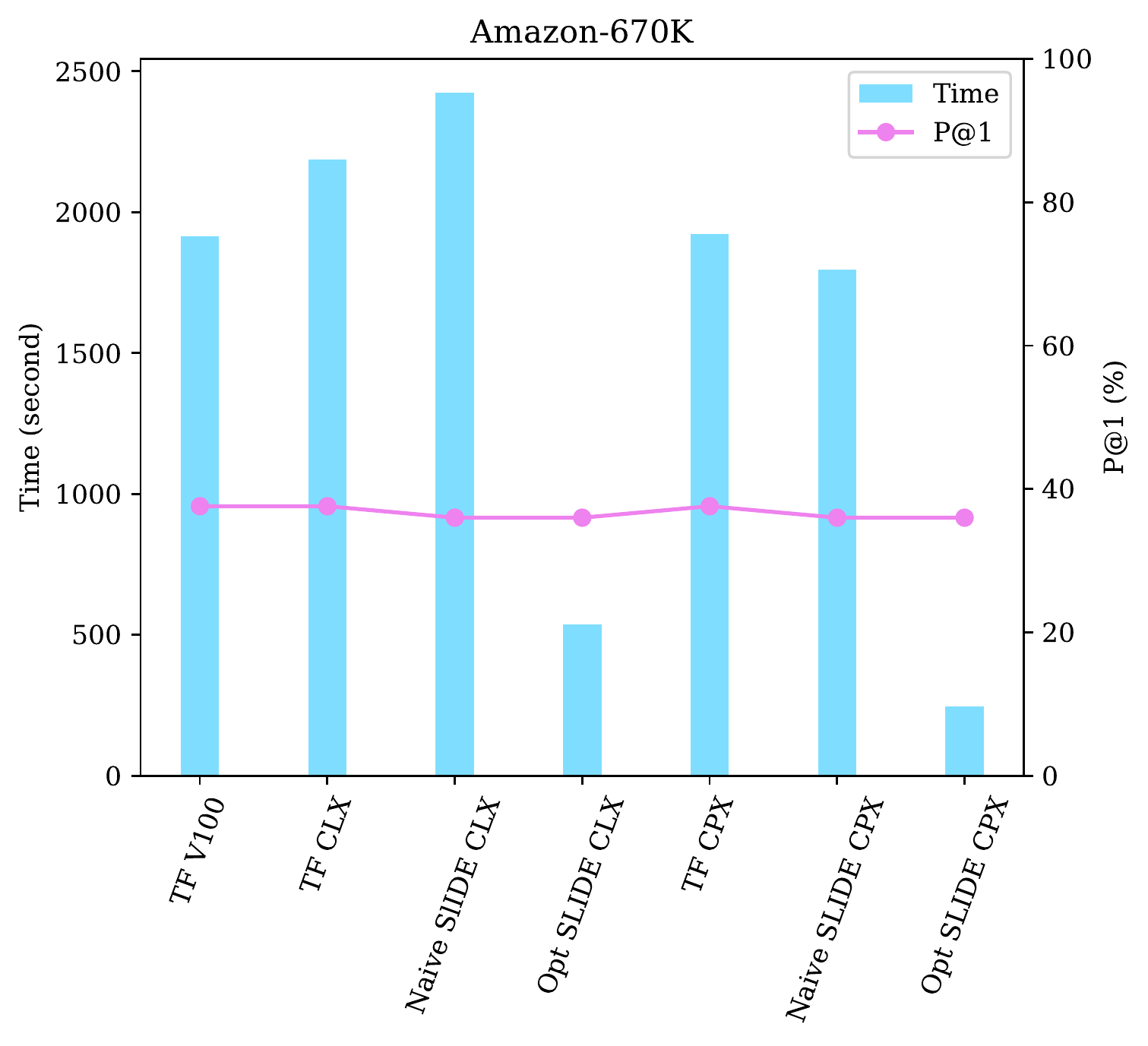}\par 
      \label{}
    \includegraphics[width=1\linewidth]{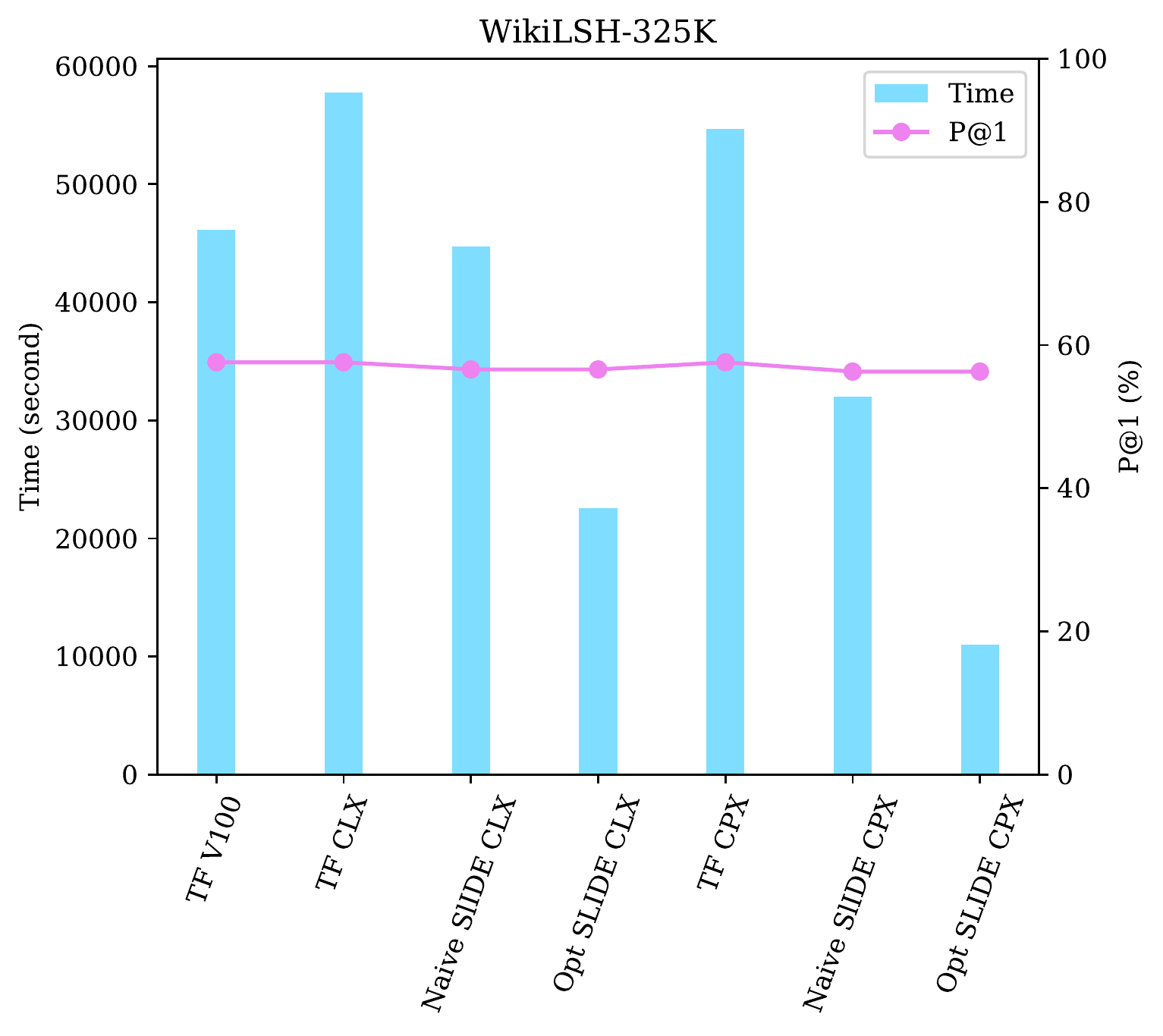}\par 
      \label{}
      \includegraphics[width=1\linewidth]{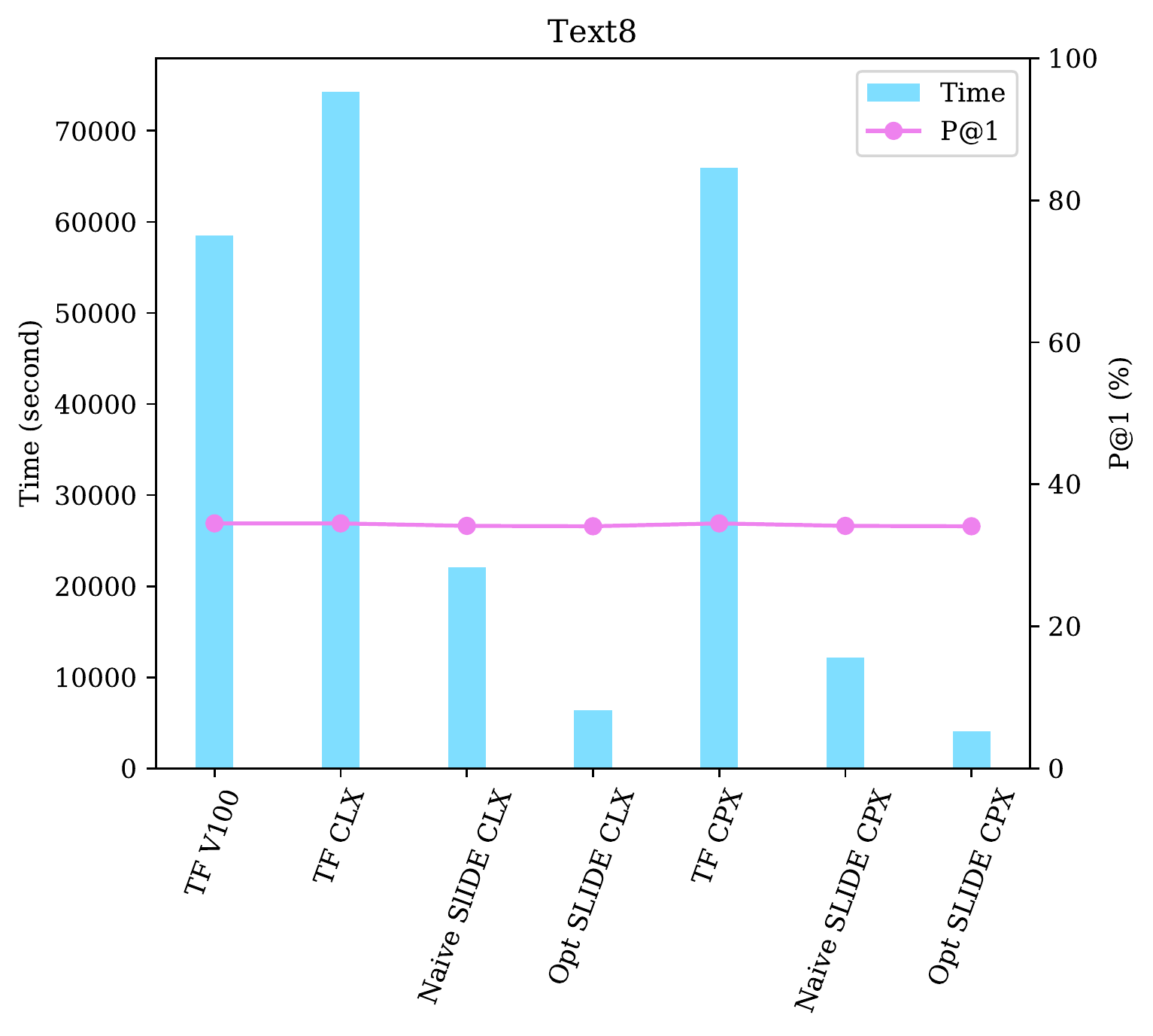}\par    
    \end{multicols}
    

\end{center}
\vspace{-0.2in}
\caption{Comparison of Optimized SLIDE on two new Intel CPUs (CLX and CPX) against the Naive SLIDE on the same CPUs and Tensorflow full-softmax on V100 and on CLX/CPX. \textbf{Top row:} Shows the convergence plots for all the methods on all datasets. The $y$-axis is \textit{Precision@1} and the $x$-axis denotes wall-clock training time in logarithmic scale. In all the convergence plots Optimized SLIDE on CPX and CLX (shown by dark and light green color respectively) outperform all other baselines with significant margin. \textbf{Bottom row:} Shows the barchart plots where the left-$y$ axis refers to average training time per epoch and right-$y$ axis refers to \textit{Precision@1}. The barchart plots confirm that Optimized SLIDE on CPX and CLX (shown as \textit{Opt SLIDE}) outperform other baselines in terms of wall-clock training time while maintain the accuracy pretty close to full-softmax in terms of \textit{Precision@1} (P@1) which is shown by pink color.}
\label{fig:main_plot}
\end{figure*}

\begin{table*}[h]
    \small
    \centering
    \caption{Average  wall-clock training time per epoch for all methods compared to TF V100. Comparison of  Optimized SLIDE against Naive SLIDE and TF-CPU on the same hardware is also reported. }
    \label{table:MainComparison}
    \begin{tabular}{|p{1.5cm}|p{1cm}|p{1.5cm}|p{1.5cm}|p{1.5cm}|p{1.5cm}|p{3.5cm}|p{3.5cm}|} \hline
    Dataset	& TF V100 & TF CLX & TF CPX & Naive SLIDE CLX & Naive SLIDE CPX &Optimized SLIDE CLX& Optimized SLIDE CPX

 \\ \hline
    Amazon-670K	 & Baseline & 1.15x slow  & 1.01x slow  & 1.26x slow & 1.1x fast& \textbf{3.5x} fast over V100  & \textbf{7.8x} fast  over V100
 \\\hline  &  &  & & & & \textbf{4x} fast over TF-CLX  & \textbf{7.9x} fast over TF-CPX
 \\\hline&  &  & & & &\textbf{4.4x} fast over SLIDE-CLX  & \textbf{7.2x} fast over SLIDE-CPX
\\
 \hline \hline
       WikiLSH-325K	 & Baseline & 1.25x slow  & 1.18x slower  &  1.03x  fast & 1.44x fast& \textbf{2.04x} fast over V100 & \textbf{4.19x} fast over V100
 \\	\hline  &  &  & & & & \textbf{2.55x} fast over TF-CLX  & \textbf{5.2x} fast over TF-CPX)
 \\\hline&  &  & & & &\textbf{2x} fast over SLIDE-CLX 
 & \textbf{3x} fast over SLIDE-CPX\\
 
 \hline \hline
       Text8	 & Baseline & 1.27x slow  &1.12x slow& 2.6x fast& 4.7x fast&   \textbf{9.2x} fast over V100 & \textbf{15.5x} fast over  V100
  \\\hline  &  &  & & & & \textbf{11.6x} fast over TF-CLX  & \textbf{17.36x} fast over TF-CPX
 \\\hline&  &  & & & &\textbf{3.5x} fast over SLIDE-CLX  & \textbf{3x} fast over SLIDE-CPX\\	 
 \hline
 \hline\end{tabular}
    	\label{table:Num_results}
\end{table*}

\section{Evaluations}
We focus on large scale evaluations where the required neural network possesses hundreds of millions of parameters. 

We compare Optimized SLIDE on two Intel's CPU machines, CPX server and CLX server, (details are in section \ref{sec:CPXCLX} ) against five baselines: 1) tensorflow implementation of full-softmax on V100 GPU 2) tensorflow implementation of full-softmax on CPX 3) tensorflow implementation of full-softmax on CLX 4) Naive SLIDE on CPX and 5) Naive SLIDE on CLX. 

It should be noted that our focus in this paper is to investigate optimizing CPU implementation of the SLIDE algorithm. We did not investigate whether the same algorithm can be applied and/or optimized for the GPU architecture, and as a result, we are reporting out-of-box results for V100 GPU.

{\bf TensorFlow Automatic Mixed Precision (AMP) does not improve these datasets:} Using both 16-bit and 32-bit floating-point numbers to accelerate computation and decrease memory space has also been employed in NVIDIA's Automatic Mixed Precision (AMP)~\cite{AMP:2020}. Although mixed-precision training has shown 1.5-3x speedup compared to FP32 training on some large models, the speedup is not always guaranteed. It requires a significant portion of operations to be converted to FP16. However, whether an operation could be converted to FP16 depends on the network structure, the types of operations ahead/afterward, etc., both for forward and backward pass. This could potentially lead to a minimal conversion ratio and slight speed improvement in a certain scenario. In our example, only less than $10\%$ of nodes are converted to BF16 due to the above reasons. There is no noticeable speed improvement by introducing AMP training (without further targeted optimization). In fact, we observe slide degradation in performance of V100. Thus, we only report the best performing version which is without the use of AMP. 

\subsection{Datasets}
We evaluate our framework and other baselines on three real public datasets. Amazon670K is a Kaggle dataset~\footnote{https://www.kaggle.com/c/extreme-classification-amazon} for recommendation systems. It was also used as evaluations in the SLIDE paper. WikiLSH-325K is a dataset taken from  ~\cite{Bhatia16} and Text8 ~\cite{matt2011} is an NLP dataset. The detailed statistics about the dimensions and samples sizes are included in Table \ref{table:data}.

\textbf{Amazon-670K} dataset is a product recommendation dataset with 670K labels. Here, each input is a vector representation of a product, and the corresponding labels are other products (among 670K choices) that a user might be interested in purchasing. This is an anonymized and aggregated behavior dataset from Amazon and poses a signiﬁcant challenge owing to a large number of classes.

\textbf{WikiLSHTC-325K} dataset is extracted from Wikipedia and consists over 1.7 Million training samples, 325K labels correspond to the category of the instance, and 1.6 Million sparse feature dimension.

\textbf{Text8} is a popular NLP dataset and a preprocessed version of the first 100 million tokens of English Wikipedia which contains 253K vocabulary. We utilize the standard word2vec language model for Text8 dataset, where input and output are one-hot and multi-hot encoded vectors, respectively. 

In word2vec architecture we utilize the skip-gram model introduced in ~\cite{mikolov2013efficient}. Skip-gram model aims to predict nearby words in a document by learning continuous representation of words. Particularly given a word, skip-gram model targets to predict the $n$ left and $n$ right neighbor words, where $n$ is window size and is considered as a hyperparameter. Window size is set to 2.

\begin{table*}[h]
    \vspace{-0.2in}
    \centering
    \caption{Impact of BF16 on average wall clock time per epoch.}
    \begin{tabular}{|p{3.5cm}|p{5.5cm}|p{3.5cm}|p{2cm}|} \hline
    Dataset	& 
 BF16 for both activations and weights & BF16 only for activations & Without BF16 

 \\ \hline
    Amazon-670K	 & Baseline & 1.16x slower & 1.28x slower 
 \\ \hline
    WikiLSH-325K	 & Baseline & 1.31x slower & 1.39x slower
\\	 \hline
      Text8	 & 2.8x slower & 1.15x faster  & Baseline
 \\	 
    	            
    	            \hline

    	\hline\end{tabular}
    	\label{table:BF16_impact}
    	\vspace{-0.2in}
\end{table*}

\subsection{Infrastructure}
\label{section:infras}
As described in Section~\ref{sec:CPXCLX}, we use two different Intel CPU servers to run our experiments: Cooper Lake server (CPX) and Cascade Lake server (CLX). 

We utilize CPX to run Optimized SLIDE with BF16, Naive SLIDE and Tensorflow implementation of full-softmax (TF-CPU 2.1) which we refer them as \textit{'Optimized SLIDE CPX'},  \textit{'Naive SLIDE CPX'} and \textit{'TF FullSoftmax CPX'}, respectively. Similarly, we use CLX to run Optimized SLIDE without BF16, Naive SLIDE and TF-CPU 2.1 and report them as \textit{'Optimized SLIDE CLX'}, \textit{'Naive SLIDE CLX'} and \textit{'TF FullSoftmax CLX'}, correspondingly. 
To run Tensorflow implementation of full-softmax on GPU, we use NVIDIA Tesla V100 Volta 32GB GPU and TF-GPU 2.1, we report it as \textit{'TF FullSoftmax V100'}. Optimized SLIDE is written in C++ and compiled under ICC with OpenMP flag and we will show that it provides significant speedup compared to non-optimized SLIDE 
in terms of wall-clock training time on all three datasets.


\subsection{Hyperparameters}
For Amazon-670K and WikiLSH-325K we use a standard fully connected neural network with hidden layer size of 128, where both the input and output are multi-hot encoded vectors. For Text8, We utilize standard word2vec language model with hidden layer size of 200, where input and output are one-hot and multi-hot encoded vectors, respectively. Effectively, we are training on the order of hundreds of million parameter neural networks on each of these datasets, where scalability is critical. See Table~\ref{table:data} for model size

We performed hyperparameter tuning for all the baselines to maintain their best trade-off between convergence time and accuracy. For all the experiments, the optimizer is Adam with a learning rate of 0.0001. The batch size for Amazon-670K, WikiLSH-325K and Text8 is 1024, 256 and 512 respectively for all the experiments. We apply hash functions for the last layer where we have the computational bottleneck. We use DWTA hash function for Amazon-670K and WikiLSH-325K with $K=6$, $L=400$ and $K=5$, $L=350$ respectively. For Text8 we use the Simhash hash function with $K=9$ and $L=50$. $L$ refers to the number of hash tables where each of the tables contains $2^K$ buckets.

{\bf A Note on Large Batch Size for Amazon-670k}
Amazon-670K dataset was the same dataset used in the SLIDE paper~\cite{chen2019slide}. In that paper, the comparisons were performed on three different batch sizes 64, 128 and 256 following standard recommendations. However, we realized that all those batch sizes only lead to around 32\% precision (even with standard TensorFlow) whereas a batch size of 1024 reaches precision of more than 35\% in a few epochs. Since we want to maintain the best accuracy, we chose the new settings for a fairer comparison. 
\begin{table}[H]
\vspace{-0.2in}
    \centering
    \caption{Impact of AVX-512 on average training time per epoch}
    \begin{tabular}{|p{2.25cm}|p{2.25cm}|p{2.5cm}|} \hline
    Dataset	& With AVX-512 & Without AVX-512 

 \\ \hline
    Amazon-670K	 & Baseline &1.22x slower
 \\ \hline
       WikiLSH-325K	 & Baseline &1.12x slower
 \\	 \hline
      Text8	 & Baseline &1.14x slower
 \\	 
    	            
    	            \hline

    	\hline\end{tabular}
    	\label{table:Avx_impact}
    	\vspace{-0.2in}
\end{table}

\subsection{Results}
We show the performance of all methods with respect to wall-clock training time and P@1.
Figure \ref{fig:main_plot} top row represents time-wise convergence plots for all datasets and show that the proposed Optimized SLIDE on both CPX and CLX (shown with dark and light green color) outperforms all other baselines with significant margin in terms of wall-clock training time. The bottom row in Figure \ref{fig:main_plot} shows the barchart for all the datasets, where the left-y label and the right-y label correspond to average training time per epoch  and P@1 respectively, and the pink line denotes P@1 value. 
We can see in all the figures that the Optimized SLIDE is much faster that all the other baselines on all the datasets consistently, while maintaining pretty similar performance to full-softmax in terms of P@1. Table \ref{table:Num_results} represents detailed numerical results for all three datasets. The first row for each dataset denotes the comparison of average training time per epoch against TF-GPU on V100. Optimized SLIDE on CPX/CLX is \textbf{7.8x/3.5x} faster than TF-GPU on V100 for Amazon-670K dataset, \textbf{4.2x/2x} faster than TF-GPU on V100 for WikiLSH-325K dataset, and \textbf{15.5x/9.2x} faster than TF-GPU on V100 for Text8 dataset. The second row of table for each dataset shows the comparison of Optimized SLIDE against TF-CPU on the same server, which shows that Optimized SLIDE on CLX/CPX is \textbf{4x/7.9x} faster than TF-CPU on CLX/CPX for Amazon-670K dataset, \textbf{2x/3x} faster than TF-CPU on CLX/CPX for WikiLSH-325K dataset, and \textbf{11.6x/17.36x} faster than TF-CPU on CLX/CPX for Text8 dataset. The third row for each dataset in the table represents the comparison of Optimized SLIDE against Naive SLIDE and TF-CPU on the same CPU server with respect to average training time per epoch. According to these rows, Optimized SLIDE on CLX/CPX is \textbf{4.4x/7.2x} faster than Naive SLIDE on CLX/CPX for Amazon-670K dataset, \textbf{2x/3x} faster than Naive SLIDE on CLX/CPX on WikiLSH-325K dataset and \textbf{3.5x/3x} faster than Naive SLIDE on CLX/CPX for Text8 dataset. These numerical results confirm the superiority of  Optimized SLIDE on CLX/CPX over all other methods on all datasets with significant margin.


    	            



    	            


\subsection{Impact of AVX-512}
To evaluate the impact of vectorization with AVX-512, we chose the same setting for all datasets in Table \ref{table:Num_results} (Optimized SLIDE on CPX). We switched off AVX-512 flag in the Optimized SLIDE code. Table \ref{table:Avx_impact} shows the results of the comparison between Optimized SLIDE with and without AVX-512 in terms of average wall-clock training time per epoch. AVX-512 vectorization decreases average training time up to 1.2x. Since it is running the same computations, the accuracy of the dataset remains unchanged.

    

    	            


\subsection{Impact of BF16}
We show the impact of BF16 on average training time per epoch in Table \ref{table:BF16_impact}. We provide three modes with respect to BF16 utilization: 1) BF16 for weights and activations 2) BF16 for activations 3) No BF16. Please refer to Section \ref{section:BF16Opt} for more details. We applied these three different modes on the best setting for each dataset (Optimized SLIDE on CPX) to evaluate the impact of BF16. According to the table, BF16 utilization for both activation and weights boosts the performance up to 1.28x and 1.39x for Amazon-670K and WikiLSH325K, respectively. However, we do not see the same phenomenon for Text8. BF16 on both activation and weights hurts the performance. BF16 only on activation boosts the performance by 1.15x compared to "No BF16," which still shows the benefit of BF16.

    


    	            


\subsection{Impact of Memory Optimizations}

Our memory optimization in Section~\ref{sec:optimize_SLIDE} provides the maximum benefits. Like other machine learning workloads, SLIDE is a memory-bound computation with a significant amount of DRAM access for loading the data and parameters. It is quite difficult to quantify the memory improvements in a randomized SLIDE where the operations are stochastically dependent on several events in the stack trace.  

However, we can still quantify the benefits of memory optimization by looking at overall speedup and discounting the speedup obtained by AVX and bfloat modifications. Our new SLIDE implementation is anywhere from 2-7x faster (Table~\ref{table:MainComparison}) than the older version. The combined speedup of AVX and bfloat instruction is around 1.7x. Thus, memory optimizations provide the remaining  performance improvement.  

\section{Conclusion}

Platform performance to power artificial intelligence need to double every two years to cope with the increase in model sizes and complexity. The performance speedup is typically achieved by both hardware and software improvements. This paper showed how merely focusing on later and less expensive \emph{software improvements} making use of widely available x86 hardware features, can provide more than double the performance. We can probably get away with the burden of handling and maintaining specialized hardware. Scaling up AI on familiar commodity hardware will increase its wider adoption, leading to AI's true democratization. 

We hope more novel software improvements will follow.  



\section{Acknowledgements}
This work was supported by National Science Foundation IIS-1652131, BIGDATA-1838177, AFOSR-YIP FA9550-18-1-0152, ONR DURIP Grant, and the ONR BRC grant on Randomized Numerical Linear Algebra.

\bibliography{example_paper,ref}
\bibliographystyle{mlsys2021}

%


\end{document}